\documentclass[a4paper,11pt]{article}
\usepackage{subcaption}
\usepackage{graphicx} 
\usepackage{xcolor}
\graphicspath{{./}}
\usepackage{amsmath}
\usepackage{amssymb}
\usepackage[thinc]{esdiff}
\usepackage{booktabs}
\usepackage{array}
\usepackage{tabularx}
\usepackage{multirow}
\usepackage{float}
\usepackage{geometry}
\title{Enhancing Histopathological Image Classification via Integrated HOG and Deep Features with Robust Noise Performance}
\author{Ifeanyi Ezuma, Ugochukwu Ugwu }
\date{}
\begin{document}
\maketitle
\begin{abstract}
\noindent The era of digital pathology has advanced histopathological examinations, making automated image analysis essential in clinical practice. This study evaluates the classification performance of machine learning and deep learning models on the LC25000 dataset, which includes five classes of histopathological images. We used the fine-tuned InceptionResNet-v2 network both as a classifier and for feature extraction. Our results show that the fine-tuned InceptionResNet-v2 achieved a classification accuracy of 96.01\% and an average AUC of 96.8\%. Models trained on deep features from InceptionResNet-v2 outperformed those using only the pre-trained network, with the Neural Network model achieving an AUC of 99.99\% and accuracy of 99.84\%. Evaluating model robustness under varying SNR conditions revealed that models using deep features exhibited greater resilience, particularly GBM and KNN. The combination of HOG and deep features showed enhanced performance, however, less so in noisy environments.
\end{abstract}

\section{Introduction}
Digital pathology has revolutionized histopathological examinations, becoming a crucial technology in contemporary clinical practice and increasingly essential within laboratory settings [1,2]. Modern histopathological images are huge, containing up to billions of pixels [3], and present complex features that are often too intricate to be represented by simple mathematical equations or models [4]. This complexity has driven the development of innovative automated methods for image analysis, facilitating more accurate and efficient diagnostics. Advances in Machine Learning (ML) and Artificial Intelligence (AI) have further enhanced the automation of these approaches, enabling the extraction of meaningful patterns from vast amounts of image data. Presently, researchers and clinicians can now leverage sophisticated algorithms and Deep Learning (DL) models to analyze complex datasets with greater precision, facilitating early disease detection and personalized treatment plans.\\

\noindent  Traditional ML techniques for histopathological image analysis typically involve several preprocessing steps, including feature selection, image segmentation, and classification [5]. These methods have been extensively reviewed in various literature as in [6][7][8]. However, over the last decade, there has been a notable shift towards the development and deployment of DL techniques to enhance and optimize complex biomedical image analysis. Specifically, after a successful application of Convolutional Neural Networks (CNNs) in the 2012 ImageNet Large Scale Visual Recognition Challenge [9 - 11], new open challenges like CAMELYON16 [12][13] for lymph node metastasis detection and PANDA [14] for Gleason grading in prostate cancer were introduced. These challenges resulted in significant advancements, demonstrating that classification systems based on CNN and Multiple-Instance Learning (MIL) could outperform pathologists in these diagnostic tasks [15]. This shift underscores the increasing relevance of DL techniques in biomedical imaging, highlighting their ability to deliver superior performance in complex diagnostic scenarios.

\section{Related works}
In recent years, pathologists have benefited from extensive new resources that have enhanced their approach to patient healthcare, consequently elevating the associated outcomes [16]. Building on this progress, researchers have carried out investigations into segmenting different structures in breast histopathological images, employing methods such as thresholding, fuzzy c-means clustering, and adaptive thresholding, attaining results that vary in their level of success [17 - 20]. Further advancements in the field include the use of Support Vector Machine (SVM) to classify skin biopsies as melanoma or nevi based on hematoxylin \& eosin (H\&E) staining. An SVC model was developed using four features and, when tested under various magnifications, performed best at 40x with the 3+4 feature combination, achieving 90\% accuracy [21]. In [22], the authors implemented various feature extraction techniques, including VGG16, InceptionV3, and ResNet50, to create eight distinct pre-trained CNN models for lung and colon cancer classification. These models achieved accuracy rates between 96\% and 100\%.\\

\noindent Additionally, the study in [23] utilized Otsu thresholding, morphological operations, and histological constraints for segmenting Whole Slide Images (WSIs). Machine Learning models, including Random Forest (RF), K-Nearest Neighbors (KNN), Support Vector Machine (SVM), and logistic regression, were trained with features extracted from segmented image patches. The Random Forest model achieved the highest accuracy at 93\%. Moreover, Yan et al. [24] developed a novel hybrid deep neural network combining convolutional and recurrent structures for classifying breast cancer histopathological images into four categories. Their method, which divided images into 12 patches for feature extraction using Inception-V3, achieved 91.3\% accuracy, outperforming current methods. The study detailed in [25] focused on classifying histopathology images of lung cancer using the LC25000 dataset and convolutional neural networks (CNNs). Feature extraction was conducted with ResNet50, VGG19, InceptionResNetV2, and DenseNet121. To improve performance, the Triplet loss function was applied. The classification task utilized a CNN with three hidden layers. Among the models, Inception-ResNetv2 achieved the highest test accuracy rate, reaching 99.7\%.

\section{Contributions}

This paper makes significant contributions through the following innovative methodologies: \\

\noindent \textbf{i. Integration of HOG and Deep Features for Enhanced Classification}: The primary contribution of this study is the introduction of a novel method that integrates Histogram of Oriented Gradients (HOG) features with deep features extracted from the InceptionResNet-v2 network. This combined feature set leverages the strengths of both traditional HOG features, which capture critical edge and texture information, and deep learning features, which provide high-level abstract representations of the images. This approach significantly improves classification accuracy compared to conventional methods. This method stands out in the literature as it offers a robust alternative to traditional methods, such as LBP and Haralick, which have shown lower performance in previous studies [26].\\

\noindent  \textbf {ii. Demonstrated Robustness and Resilience}: The study also evaluates the robustness of the proposed models under varying conditions, particularly in terms of noise and signal-to-noise ratio (SNR). The models trained on deep features exhibited greater resilience to these variations, ensuring high performance even in less-than-ideal imaging conditions. The findings suggest that the proposed integrated method not only enhances accuracy but also maintains high reliability, which is contrary to existing approaches in this field [27].

\section{Material and Methods}
Each experiment involved utilizing the provided architecture of InceptionResNet-V2. By leveraging this pre-trained network, we evaluated its performance in two distinct scenarios: first, as a standalone classifier and second, as a feature extractor. In the classification task, we directly applied InceptionResNet-V2 to categorize the images. For feature extraction, we used the network to obtain high-level features from the images, which were then fed into additional machine learning models for the final classification.

\subsection{Dataset}
The LC25000 dataset contains 25,000 de-identified, HIPAA-compliant, and validated color images [28]. It comprises five classes: colon adenocarcinoma, benign colonic tissue, lung adenocarcinoma, lung squamous cell carcinoma, and benign lung tissue, each containing 5,000 images. The images were initially cropped to 768 x 768 pixels before enhancements such as left and right rotations (up to 25 degrees, with a 1.0 probability) and horizontal and vertical flips (with a 0.5 probability) were applied [29]. The data was then split into 60\% for training, 20\% for validation, and 20\% for testing. Figure 1 shows samples from each class in the dataset.\\

\begin{figure}[ht]
    \centering
    \includegraphics[width=0.65\textwidth]{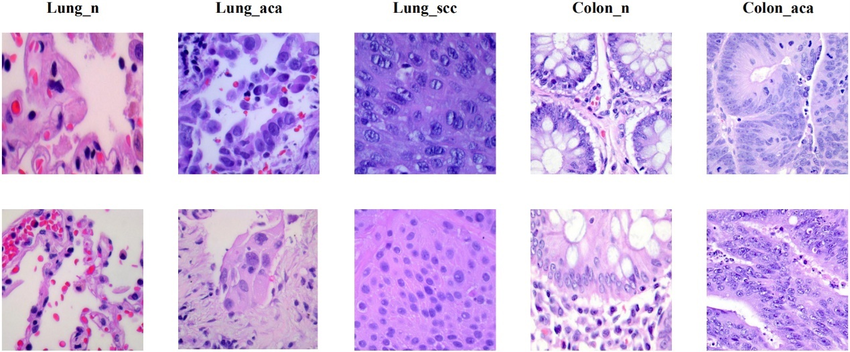} 
    \caption{Dataset LC25000. From left to right: benign lung tissue (Lung\_n), lung adenocarcinoma (Lung\_aca), lung squamous cell carcinoma tissue (Lung\_scc), benign colonic tissue (Colon\_n) and colon adenocarcinoma (Colon\_aca) [30].} 
    \label{fig:model}
\end{figure}

\subsection{Fine-Tuned Pre-trained CNN as Classifier}
During the fine-tuning process, two convolutional blocks were defined, each consisting of a convolutional layer, a batch normalization layer, a ReLU activation function, and a max pooling layer. The first convolutional layer is composed of 128 kernels of size 7x7, with padding set to 3 to preserve the spatial dimensions of the input image after the convolution operation. Subsequently, the second convolutional layer employs 256 filters of size 5x5, with the stride set to 2, which effectively reduces the feature map dimensions by a factor of 2. Following these layers, it's a fully connected layer to integrate the learned features for the final classification task. For optimization, we implemented Stochastic Gradient Descent with a Momentum value of 0.9. The initial learning rate was set at $10^{-4}$, and a piecewise learning rate schedule was adopted to determine how the learning rate changes over epochs.

\subsection{Pre-Trained CNN for Feature Extraction}
The images were augmented through random rotations (0°, 90°, 180°, 270°) and translations ([-3, 3] pixels) to increase the diversity of the training set. Features were extracted from the augmented images using the 'avg\_pool' layer of the InceptionResNet-v2 network. Several ML models were then trained on these features using Bayesian optimization for hyperparameter tuning.

\subsection{Integration  of Hog Features with Features Extracted from Pre-Trained CNN}
The images were initially cropped to a size of 299 x 299 pixels and augmented through random rotations and translations. HOG features were extracted using a cell size of 128 x 128 pixels, and deep features were obtained from the 'avg\_pool' layer of the InceptionResNet-v2 network. These features were then concatenated to form a comprehensive feature vector. The same ML models  as in section 4.2 were then trained on these combined features, with hyperparameters optimized through Bayesian optimization.

\section{Results}

The learning rate and training loss, depicted in Figure 2(a), demonstrate the model’s rapid adaptation to the specific features of the histopathological images, with the training loss stabilizing as the optimal learning rate is achieved. Minor fluctuations in the training loss indicates \\

\begin{figure}[H]
    \centering
    \begin{subfigure}[b]{0.35\textwidth} 
        \vspace{-20mm} 
        \centering
        \raisebox{7mm}{\includegraphics[width=\textwidth]{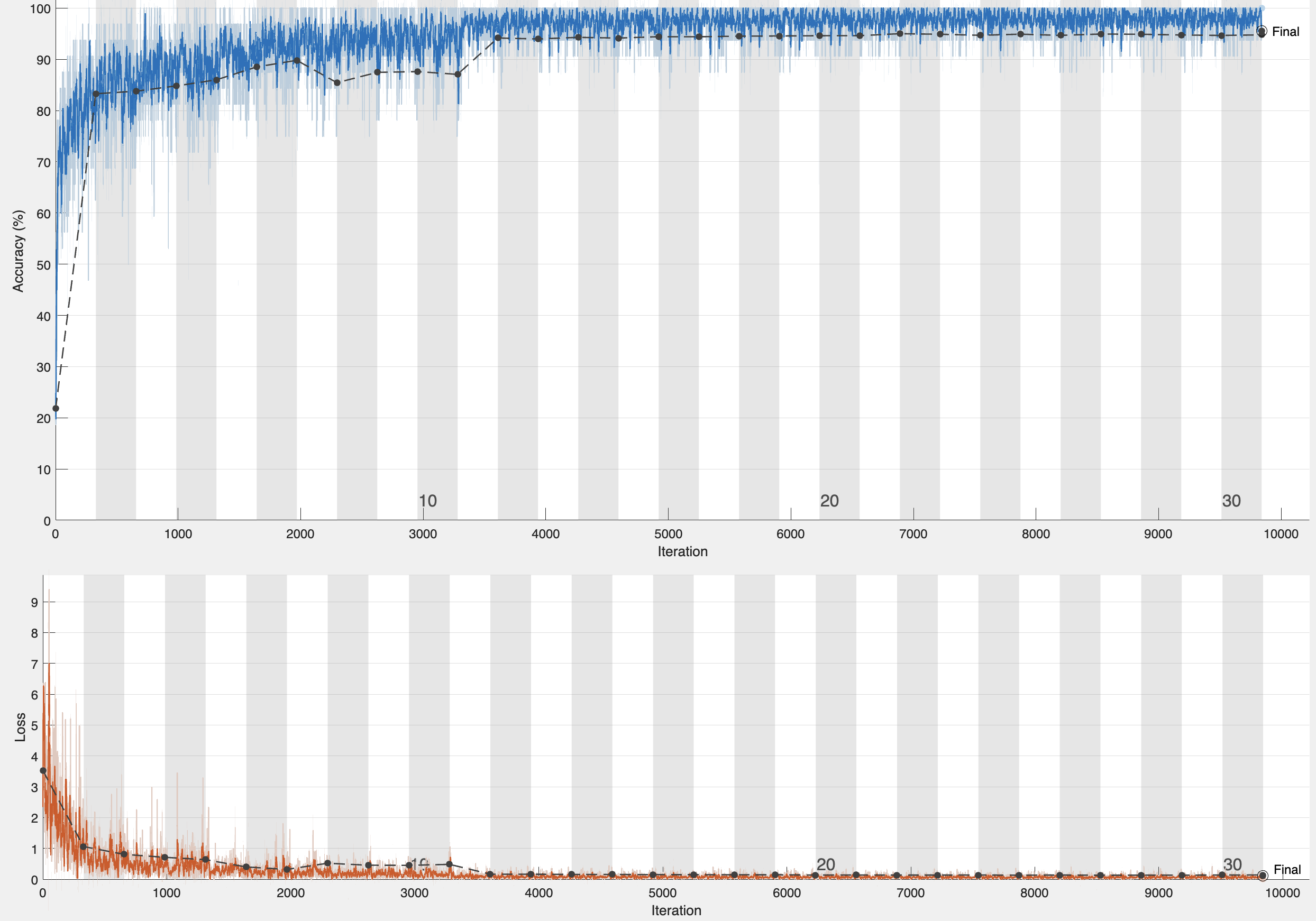}} 
        \caption*{(a)}
        \label{fig:part1a}
    \end{subfigure}%
    \hspace{0.001\textwidth} 
    \begin{subfigure}[b]{0.35\textwidth} 
        \vspace{-2mm} 
        \centering
        \raisebox{-15mm}{\includegraphics[width=\textwidth]{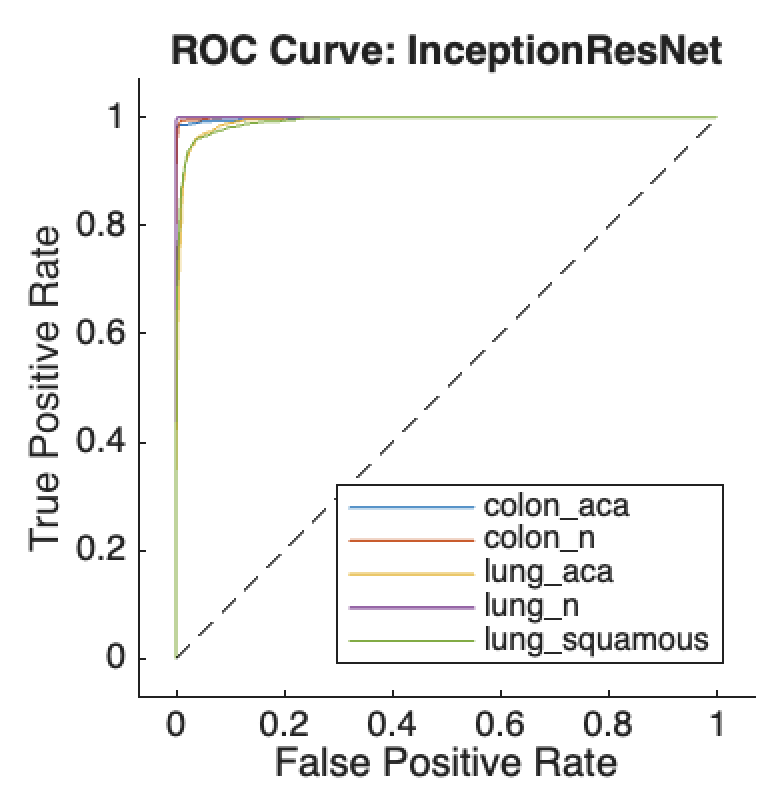}} 
        \caption*{(b)}
        \label{fig:part2a}
    \end{subfigure}
    \captionsetup{justification= justified} 
    \caption{Performance Evaluation using the Fine-tuned Pre-trained InceptionResNet Model.}
    \label{fig:model_components_a}
\end{figure}

\noindent ongoing adjustments to the dataset’s intricate features, reflecting a dynamic learning process aimed at minimizing classification errors. The ROC curves for all five classes of histopathological images, shown in Figure 2(b), exhibit high true positive rates and low false positive rates across all classes. The model achieved a notable classification accuracy of 96.01\%, with an average AUC of 96.8\%, indicating strong overall performance. \\

\begin{figure}[ht]
    \centering
    \begin{subfigure}[b]{0.37\textwidth} 
        \centering
        \includegraphics[width=\textwidth]{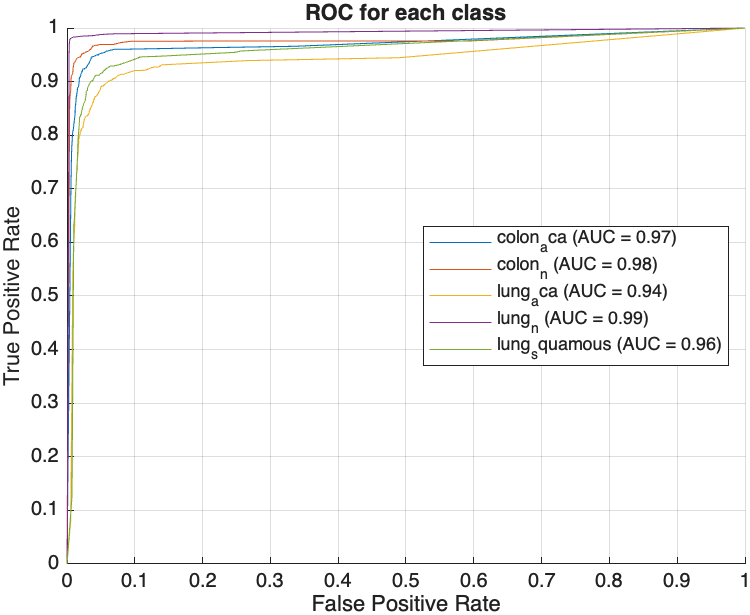} 
        \caption*{(a)}
        \label{fig:part1b}
    \end{subfigure}%
    \hspace{0.005\textwidth} 
    \begin{subfigure}[b]{0.42\textwidth} 
        \centering
        \includegraphics[width=\textwidth]{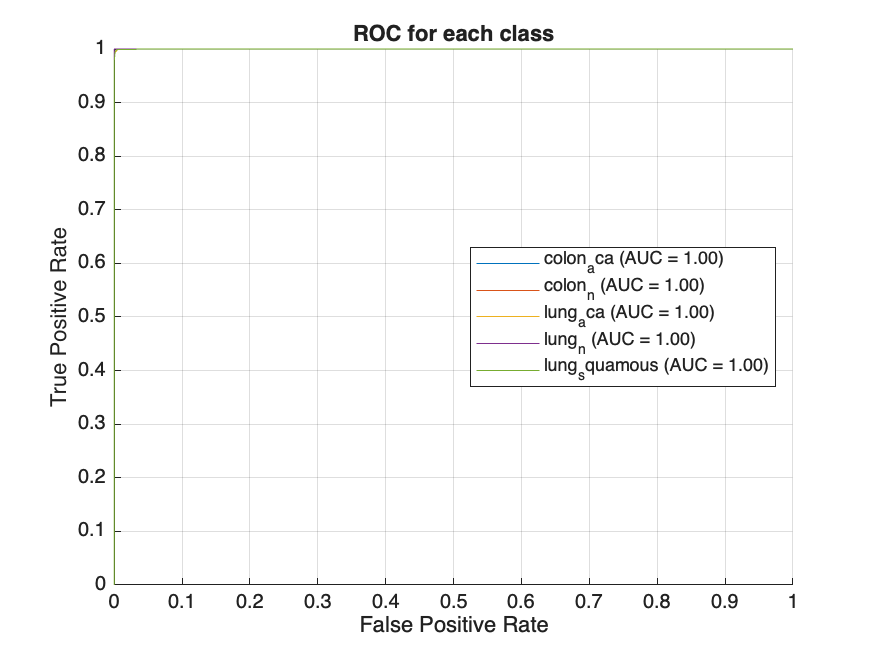}
        \caption*{(b)}
        \label{fig:part2b}
    \end{subfigure}
    \caption{ ROC curves obtained by the models. From left to right, the worst and the best performing model (a) Feature extraction using Decision Tree (b) Feature extraction using Neural Network.}
    \label{fig:model_components_b}
\end{figure}

\noindent Following feature extraction, several machine learning models were trained on the augmented dataset, with hyperparameter tuning conducted via Bayesian optimization. The models evaluated included Decision Tree, Gradient Boosting Machine (GBM), k-Nearest Neighbors (KNN), Neural Network, and Support Vector Machine (SVM). The performance of these models was assessed based on two key metrics: AUC and test accuracy. The ROC curves obtained by the models are shown in Figure 3, with plots arranged from left to right, representing the worst to the best performing model. The Decision Tree model, illustrated in Figure 3(a), achieved an AUC of 95.85\% and a test accuracy of 91.36\%. While this indicates a solid performance, it is relatively lower compared to the other models evaluated. Significant improvements were observed with the GBM model, which achieved an AUC of 99.77\% and a test accuracy of 98.05\%. KNN, SVM, and Neural Network models demonstrated near-perfect performance, each achieving an AUC and test accuracies of over 99\% as shown in Table 1. The Neural Network model, depicted in Figure 3(b), represents the best performing model in this experiment with an AUC of 99.99\% and accuracy of 99.72\%. \\

\begin{table}[h!]
\centering
\begin{tabular}{lcc}
\toprule
Model & AUC(\%) & Accuracy(\%) \\
\midrule
Decision Tree & 95.85 & 91.36 \\
GBM & 99.77 & 98.05 \\
KNN & 99.72 & 99.67 \\
Neural network & 99.99 & 99.72 \\
SVM & 99.94 & 99.03 \\
\bottomrule
\end{tabular}
\caption{Performance Metrics of Machine Learning Models on deep features extracted from InceptionResNet-V2.}
\label{tab:model_performance}
\end{table}

\noindent The study also evaluated the performance of the different machine learning models using HOG features combined with deep features extracted from the pre-trained InceptionResNet-v2 network. When compared to the models trained using only deep features from InceptionResNet-v2, there was a noticeable improvement in performance across most models with the addition of HOG features as can be seen from Table 2. Among the models, the Decision Tree exhibited a

\begin{table}[h!]
\centering
\begin{tabular}{lcc}
\toprule
Model & AUC(\%) & Accuracy(\%) \\
\midrule
Decision Tree & 95.04 & 93.83 \\
GBM & 99.9 & 98.73 \\
KNN & 99.82 & 99.8 \\
Neural network & 99.99 & 99.84 \\
SVM & 99.96 & 99.40 \\
\bottomrule
\end{tabular}
\caption{Performance Metrics of Machine Learning Models on HOG and deep features extracted from InceptionResNet-V2.}
\label{tab:model_performance_b}
\end{table}

\noindent solid but relatively modest performance, achieving an AUC of 95.04\% and a test accuracy of 93.83\%. In contrast, the GBM model demonstrated significant improvement with an AUC of 99.9\% and a test accuracy of 98.73\%. KNN and SVM models performed exceptionally well, with KNN achieving an AUC of 99.82\% and an accuracy of 99.8\%, while SVM attained an AUC of 99.96\% and an accuracy of 99.40\%. The Neural Network model emerged as the top performer, reaching an almost perfect AUC of 99.99\% and a test accuracy of 99.84\%. This suggests that combining HOG features with deep learning features enhances better classification performance.\\

\subsection{Evaluating the Impact of Signal-to-Noise Ratio (SNR) on Model Performance}
This section presents an in-depth evaluation of the impact of Signal-to-Noise Ratio (SNR) on the performance of all models that have been introduced in this study. To systematically assess the impact of SNR, we introduced different levels of noise to the dataset by applying additive Gaussian noise to the images at various SNR levels to simulate different noise conditions. The SNR levels were set at 40dB, 35dB, and 30dB to represent low, moderate, and high noise environments, respectively as shown in figure 4 below. These noisy datasets were then used to evaluate the robustness and adaptability of the models trained using the different methods described in section 3.2, 3.3 and 3.4. The impact of SNR on model performance was systematically compared across the three methods to evaluate the robustness and reliability of each model.
\begin{figure}[ht]
    \centering
    \setlength{\tabcolsep}{2pt} 
    \begin{tabular}{>{\centering\arraybackslash} m{0.17\textwidth} 
                    >{\centering\arraybackslash} m{0.17\textwidth} 
                    >{\centering\arraybackslash} m{0.17\textwidth}}
        \scriptsize 40dB & \scriptsize 35dB & \scriptsize 30dB \\
        \includegraphics[width=\linewidth]{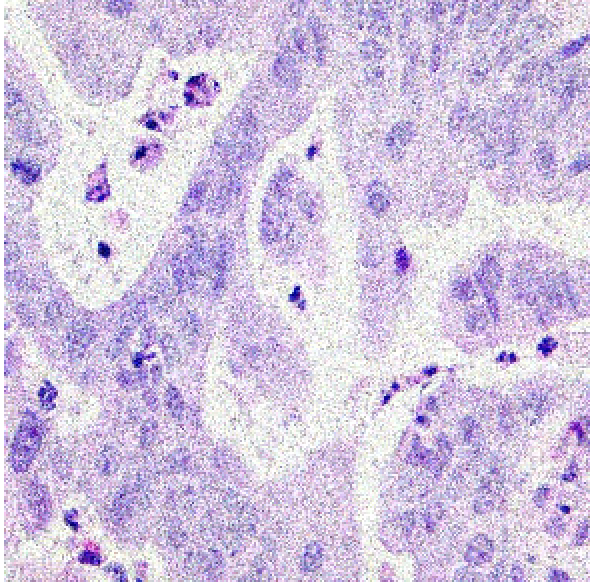} & 
        \includegraphics[width=\linewidth]{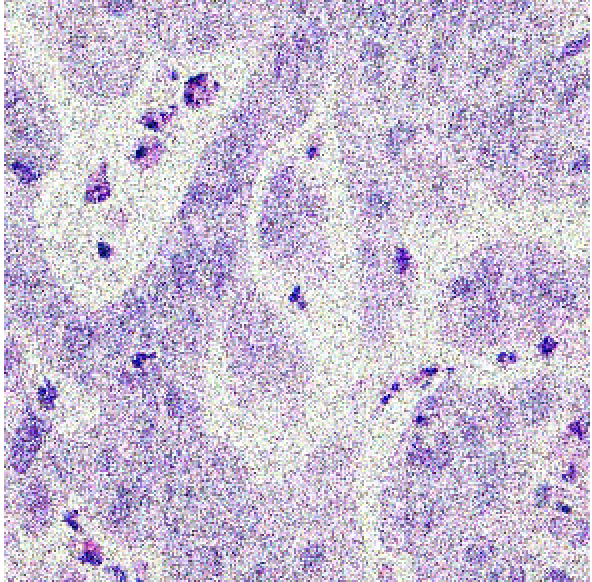} &
        \includegraphics[width=\linewidth]{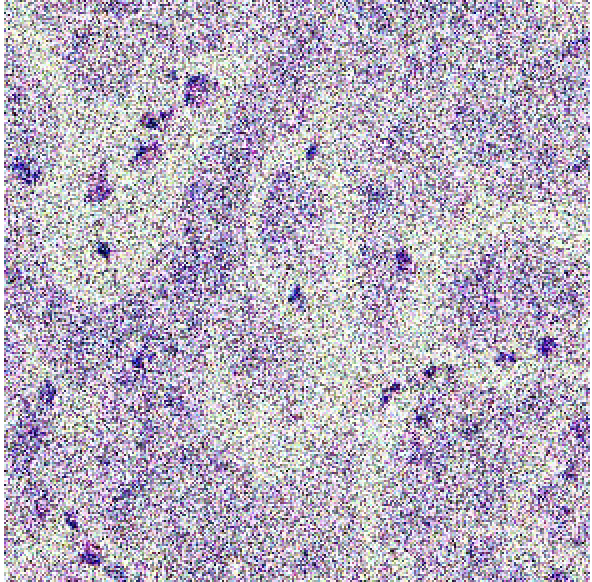}  \\
    \end{tabular}
    \caption{Images with different levels of SNR applied. From left to right representing low, moderate and high level of background noise.}
    \label{fig:snr_images3}
\end{figure}

\noindent For the InceptionResNet-V2 used directly as a classifier, performance significantly deteriorated as noise levels increased, with accuracy dropping from 80.16\% at 30dB to 27.44\% at 40dB, as shown in Table 1. In contrast, models using deep features extracted from InceptionResNet-V2 exhibited more resilient performance under varying noise conditions. As shown in Figure 2(a), \\

\begin{table}[H]
\centering
\begin{tabular}{lcc}
\toprule
Method & SNR(dB) & Accuracy(\%) \\
\midrule
\multirow{3}{*}{InceptionResNet-V2 as a classifier} & 30 & 27.44 \\ 
& 35 & 65.71 \\
& 40 & 80.16 \\
\bottomrule
\end{tabular}
\caption{Accuracy of InceptionResNet-V2 as a Classifier at Different SNR Levels.}
\label{tab:snr_40_performance}
\end{table}

\noindent the Decision Tree model's accuracy improved from approximately 50\% at 30dB to about 75\% at 40dB, while more sophisticated models like GBM, KNN, Neural Network, and SVM achieved accuracies above 85\% at 40dB. However, at 30dB, GBM attained an accuracy above 80\%, with KNN following closely at 75.04\%, while the Neural Network and SVM achieved significantly lower accuracies of 71.65\% and 66.53\%, respectively. On the other hand, with the combination of HOG features with deep features, as illustrated in Figure 2(b), all models attained similar levels of accuracy at 40dB compared with the performance of the models using deep feature extraction alone. Furthermore, there was little improvement observed with the Decision Tree \\

\begin{figure}[H]
    \centering
    \begin{subfigure}[b]{0.44\textwidth} 
        \centering
        \includegraphics[width=\textwidth]{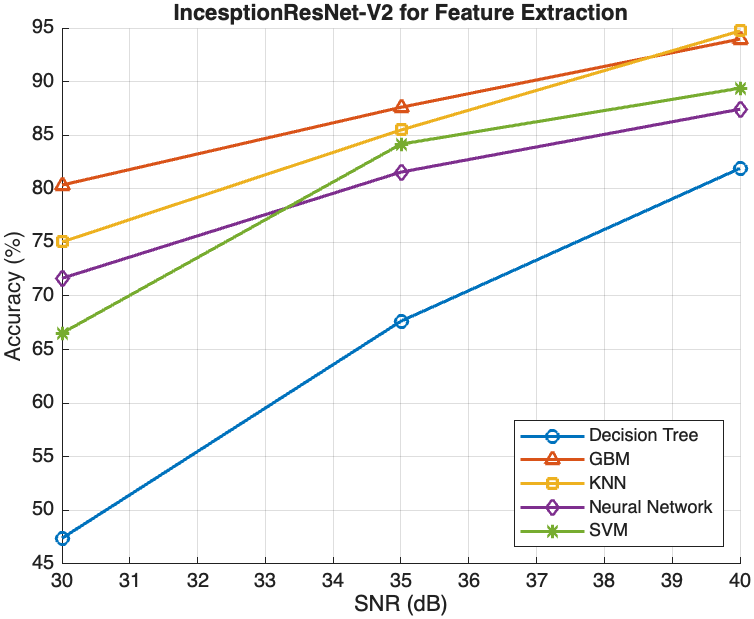} 
        \caption*{(a)}
        \label{fig:part1c}
    \end{subfigure}%
    \hspace{0.005\textwidth} 
    \begin{subfigure}[b]{0.46\textwidth} 
        \centering
        \includegraphics[width=\textwidth]{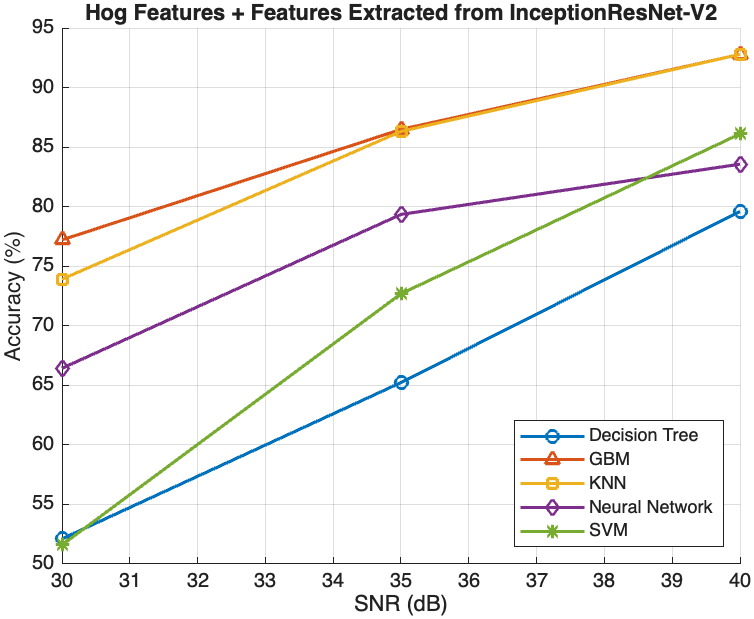}
        \caption*{(b)}
        \label{fig:part2c}
    \end{subfigure}
    \caption{Accuracy of Various Machine Learning Models Across Different SNR Levels Under different methods (a) Features extracted from InceptionResNet-V2  and (b) Combined HOG and Deep Features extracted from InceptionResNet-V2.}
    \label{fig:model_components_C}
\end{figure}

\noindent model, which achieved accuracies of 65.23\% and 52.11\% at 35dB and 30dB, respectively. However, the other models experienced relatively lower accuracies at 35dB and 30dB noise levels compared with the deep feature extraction method. GBM performed the best, achieving accuracies of 86.49\% at 35dB and 77.21\% at 30dB, followed closely by KNN with accuracies of 86.42\% and 73.21\%, respectively. Moreover, while the Neural Network and SVM models also improved with higher SNR levels, their performance gains were less pronounced compared to GBM and KNN. At 35dB and 30dB, the Neural Network model's accuracy was 79.35\% and 66.43\%, respectively, whereas the SVM model's accuracy was 72.69\% and 51.59\%, respectively.

\section{Conclusion}
This study evaluated the effectiveness of various ML and DL models in classifying histopathological images, focusing on feature extraction techniques and the impact of SNR on model performance. The fine-tuned InceptionResNet-v2 network achieved notable classification accuracy (96.01\%) and AUC (96.8\%). Models trained on deep features from InceptionResNet-v2 outperformed those using only the pre-trained network, with the Neural Network model achieving an almost perfect AUC of 99.99\% and accuracy of 99.84\%. The addition of HOG features further enhanced performance, particularly for advanced models like GBM, KNN, Neural Network, and SVM.\\

\noindent Evaluating the impact of SNR revealed that models using deep features were more resilient to noise, with GBM and KNN maintaining high accuracy even at lower SNR levels. For instance, GBM achieved above 80\% even at 30dB, while KNN maintained around 75\% accuracy under the same conditions. The combination of HOG and deep features showed similar performance trends, though with slightly lower accuracy compared to using deep features alone. For example, the Neural Network model achieved 79.35\% accuracy at 35dB and 66.43\% at 30dB with the combined features. The Neural Network model consistently emerged as the top performer across different feature extraction methods and noise levels, demonstrating the strength of combining fine-tuned neural networks with robust feature extraction techniques. Other models like GBM and KNN also showed strong performance, particularly in noisy conditions, making them suitable alternatives depending on the specific diagnostic task requirements. Overall, the integration of advanced feature extraction methods and robust machine learning models significantly enhances the classification performance of histopathological images, even under challenging conditions.

\section*{}

\end{document}